\documentclass[times,art10,twocolumn,latex8,8pt]{article}
\usepackage[top=30truemm,bottom=25truemm,left=20truemm,right=20truemm]{geometry}
\usepackage{authblk}
\usepackage{amsmath}
\usepackage{algorithm,algorithmic}
\usepackage{subcaption}
\usepackage[pdftex]{graphicx}

\title{\bf Learning Mobile CNN Feature Extraction \\ \bf Toward Fast 
Computation of Visual Object Tracking}
\author[1]{\normalsize\bf Tsubasa Murate}
\author[1]{\normalsize\bf Takashi Watanabe}
\author[1]{\normalsize\bf Masaki Yamada}

\affil[1]{\normalsize Graduate School of Infomatics, Nagoya University}

\date{}

\begin{document}
\maketitle

\begin{center}
\section*{Abstract}
\end{center}
{\it
In this paper, we construct a lightweight, high-precision and high-speed object tracking using a trained CNN.
Conventional methods\cite{FCNT}\cite{yamada} with trained CNNs use VGG16 network\cite{VGGNet} which requires powerful computational resources.
Therefore, there is a problem that it is difficult to apply in low computation resources environments.
To solve this problem, we use MobileNetV3\cite{2019movilenetv3}, which is a CNN for mobile terminals.
Based on Feature Map Selection Tracking\cite{yamada}, we propose a new architecture that extracts effective features of MobileNet for object tracking. 
The architecture requires no online learning but only offline learning.
In addition, by using features of objects other than tracking target, the features of tracking target are extracted more efficiently.
We measure the tracking accuracy with Visual Tracker Benchmark \cite{OTB} and confirm that the proposed method can perform high-precision and high-speed calculation even in low computation resource environments.
}

\section{Introduction}
Visual object tracking is a task of estimating the existence area of a object in a movie.
It is a useful technology that can be applied to various scenes including automatic vehicle technology and security cameras.
One approach of object tracking is a method using general object recognition CNN.
It is known that the features of general object recognition CNN, which contains highly accurate results in image classification, can be applied to tasks other than image classification such as segmentation.
The tracking approach applies the features of general object recognition CNN to tracking.

Feature Map Selection Tracking(FMST) \cite{yamada} is one of the tracking methods using general object recognition CNN.
In this method, the features for a tracking target are extracted from pre-trained VGG16 and the existence area of the target is predicted.
It does not need offline learning and online learning, and realizes high-precision and high-speed object tracking.
However, since VGG16 requires a huge amount of computing resources such as GPU, there is a problem that it is difficult to apply in low computation resources environments.
There are many small devices such as security cameras that are expected to be applied to object tracking technology.
For them, a method with a large network is undesirable for these technology.
In this study, we propose a lightweight, high-precision and high-speed tracking method based on the approach of FMST.

\section{Feature Map Selection Tracking}

This chapter outlines our Feature Map Selection Tracking (FMST)\cite{yamada}

\subsection{Feature Map Selection}\label{sec:2.1}
\begin{figure}[h]
  \centering
  \includegraphics[width=8cm]{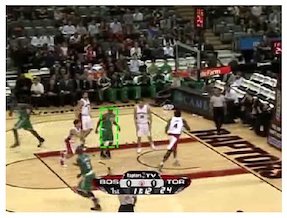}
  \caption[]{Input image (green rectangle is the target).}
  \label{fig:chap2_inputimage}
\end{figure}

In the benchmark for visual object tracking evaluation, the image and the area of the tracking target are given at the first frame.
Fig.\ref{fig:chap2_inputimage} shows the initial frame of the tracking task "Basketball"\cite{OTB}. The tracked object is visualized with a green rectangle, but in fact only the coordinates of the rectangle is given.

\begin{figure}[htbp]
 \begin{minipage}{0.43\hsize}
  \begin{center}
   \includegraphics[height=\hsize]{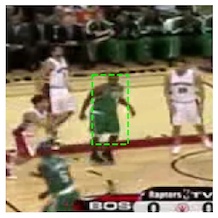}
   \caption[]{ROI.}
   \label{fig:chap2_ROI}
  \end{center}
 \end{minipage}
 \begin{minipage}{0.43\hsize}
  \begin{center}
   \includegraphics[height=\hsize]{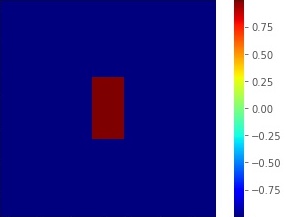}
   \caption[]{Target map.}
   \label{fig:TargetMap}
  \end{center}
 \end{minipage}
\end{figure}

The area where the tracking target exists is represented by a rectangle $\mathbf{x}=[x, y, w, h]$.
$x, y, w, h$ represent the center coordinates, width, and height, respectively.
First, centering on the tracking target, we cut out a rectangle that is twice as large as $\mathbf{x}$ in both of the vertical and horizontal directions. 
This rectangle is called Region of Interest(ROI).
Target map $\mathbf{M}$ is a matrix that has the size same as that of the ROI and its elements are $1$ in the area corresponding to $\mathbf{x}$ and $-1$ otherwise.
Figs.\ref{fig:chap2_inputimage} to \ref{fig:TargetMap} show examples of input image, ROI and target map for tracking task "Basketball", respectively.

The feature map of the $l$ th layer obtained by inputting the ROI in the trained CNN is $\mathbf{F}^l$, and the $c$ channel is expressed as $\mathbf{F}_c^l$.
Let $N_c^l$ denote the number of the channels of the $l$ th layer.
The target map $\mathbf{M}^l$ is resized with the size of the feature map of the $l$ th layer and the score $s_c^l$ of each feature map $\mathbf{F}_c^l$ is defined by the following equation:

\begin{equation}
  s^l_c = \mathrm{sum}(\mathbf{F}^l_c \circ \mathbf{M}^l)\label{eq:2-1}
\end{equation}

The $\circ$ represents the Hadamard product, $\mathrm{sum}$ represents the sum of all the elements of the matrix.
The vector with elements $s^l_c$ is defined as a score vector $\mathbf{s}^l$.

The prediction map $\hat{\mathbf{M}^l}$ is obtained by following equation.

\begin{equation}
\hat{\mathbf{M}}^l = \sum_{c\in{C_{sel}^l}}\mathbf{F}^l_c\label{eq:2-2}
\end{equation}

$C_{sel}^l$ represents a set of the channels at top $10 \%$ score.

\begin{figure}[htbp]
 \begin{minipage}{0.43\hsize}
  \begin{center}
   \includegraphics[height=\hsize]{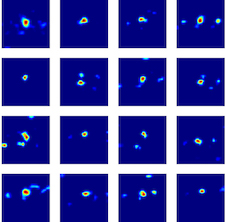}
   \subcaption[]{Top 16 feature maps}
  \end{center}
  \label{fig:3-1best16}
 \end{minipage}
 \begin{minipage}{0.43\hsize}
  \begin{center}
   \includegraphics[height=\hsize]{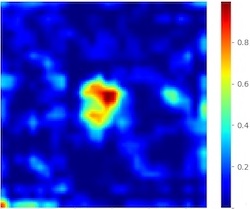}
   \subcaption[]{Prediction map $\hat{\mathbf{M}^l}$}
  \end{center}
  \label{fig:3-1sumed_fmap}
 \end{minipage}
 \caption[]{Examples of feature maps and $\hat{\mathbf{M}^l}$.}
\end{figure}

\subsection{Tracking method}

\subsubsection{Feature map average score}
In order to obtain a prediction map from feature maps $\mathbf{F}_c^l,$ where $c \in \{1,2, ... ,N_c^l\}$, the weighted sum of these feature maps is calculated.
The weight vector for $\mathbf{F}^l$ is represented by  $\mathbf{w}^l$, and its element is expressed by $w_c^l$.
The prediction map $\hat{\mathbf{M}}^l$ is evaluated by the following equation:

\begin{align}
\hat{\mathbf{M}}^l &= \mathbf{F}^l\mathbf{w}^l \\
&=\sum_{c=1}^{N^l_c}w^l_c\mathbf{F}^l_c\label{eq:2-3}
\end{align}

Each element $w_c^l$ of the weight vector $\mathbf{w}^l$ is defined as follows:

\begin{equation}
w^l_c \gets \begin{cases}
            1 & (\textrm{if}~~c \in C^l_{\textrm{sel}}) \\
            0 & (\textrm{otherwise})
            \end{cases}\label{eq:2-4}
\end{equation}

The $C_{sel}^l$ represents the channel set of the top 10 \% average score $\bar{s}^l_c$.
The $\bar{s}^l_c$ is the average of the scores over time.
The $\bar{s}^l_c$ is initialized with $\bar{s}^l_c \gets s^l_c$ in the first frame.
In the subequent frames, $\bar{s}^l_c$ is updated as shown in the following equation using the score $s^l_c$ obtained at each frame:

\begin{equation}
\bar{s}^l_c \gets \eta\bar{s}^l_c + (1-\eta)s^l_c \label{eq:2-5}
\end{equation}

\begin{figure}[h]
  \centering
  \includegraphics[width=8cm]{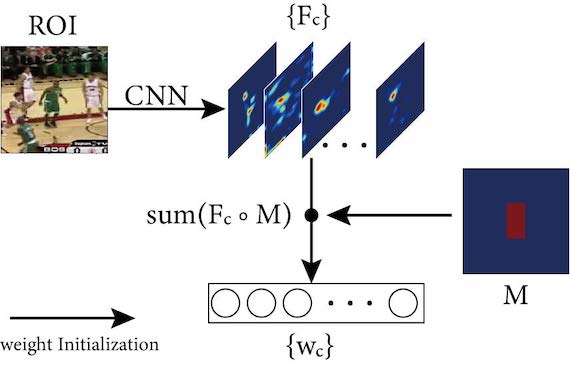}
  \caption[]{Weights initialization.}
  \label{fig:chap2_init}
\end{figure}

\begin{figure}[h]
  \centering
  \includegraphics[width=8cm]{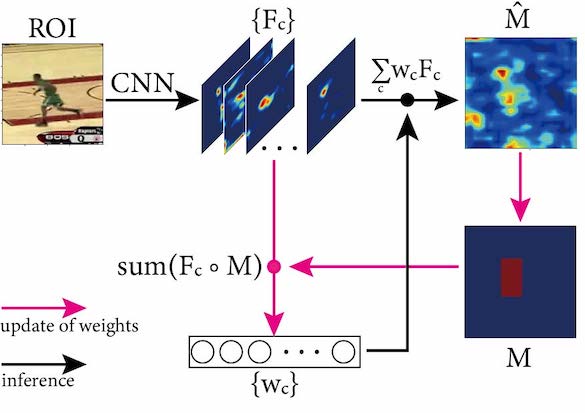}
  \caption[]{Weights update.}
  \label{fig:chap2_fmst}
\end{figure}

The $\eta$ is the smoothing coefficient, and we set $\eta = 0.99$.
By selecting the feature map using the average score, it is possible to reflect the score over time.
Therefore, even if the score of the feature map at a certain time cannot be calculated correctly due to occlusion, it is expected that it will be possible to track the object based on the score in the time afterward.
Fig.\ref{fig:chap2_init} shows the flow of weight vector initialization. Fig.\ref{fig:chap2_fmst} shows the update of weight vector.

\subsubsection{Two types of target maps}
We use two types of the target maps with different area of the element value $-1$.
Let name the two types of the target maps as Type C and Type S.
Type C and Type S target maps are represented as $\mathbf{M}^l_{\textrm{C}}$ and $\mathbf{M}^l_{\textrm{S}}$, respctively.
$\mathbf{M}^l_{\textrm{C}}$ is a matrix with the size of ROI and has its elements of $1$ in the area corresponding to $\mathbf{x}$ and $-1$ in all other areas.
The prediction map generated by using $\mathbf{M}^l_{\textrm{C}}$ is represented as $\hat{\mathbf{M}}^l_{\textrm{C}}$.

$\mathbf{M}^l_{\textrm{S}}$ also has the size of ROI, and its elements are $1$ in the area of $\mathbf{x}$, $-1$ in the area obtained by doubling the area of $\mathbf{x}$ vertically and horizontally and $0$ otherwise.
The prediction map generated by $\mathbf{M}^l_{\textrm{S}}$ is expressed as $\hat{\mathbf{M}}^l_{\textrm{S}}$.

\begin{figure}[h]
  \centering
  \includegraphics[width=8cm]{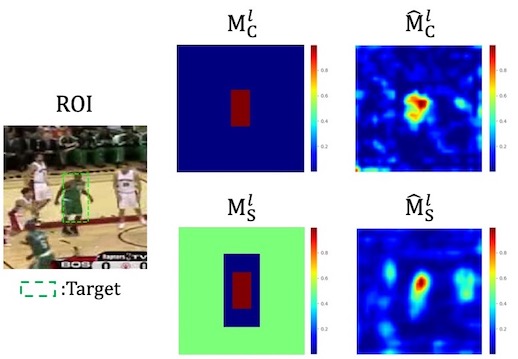}
  \caption[]{Examples of ROI, target maps and prediction maps.}
  \label{fig:tmap_sc}
\end{figure}

When the $\mathbf{M}^l_{\textrm{C}}$ is used, feature maps with strong activity only for the tracking target are selected.
Therefore, a prediction map that is effective in distinguishing the target object from other objects can be obtained.

On the other hand, since the $\mathbf{M}^l_{\textrm{S}}$ focuses only on the area very close to the tracking target, feature maps that separate the background and the object are selected. 

Fig.\ref{fig:tmap_sc} shows a comparison of the prediction maps obtained from the two types of target maps.
$\hat{\mathbf{M}}^l_{\textrm{C}}$ is able to obtain effective activity to distinguish the target from other players. In the case of $\hat{\mathbf{M}}^l_{\textrm{S}}$, it is possible to make a stronger distinction between the target object and the background.

\subsubsection{Generation of candidate areas}\label{chap2:candidate}
The rectangle $\mathbf{x}$ that surrounds the tracking target determines the candidate area for the tracking target at time $t$.
Random sampling of the target region is performed based on the method of Wang et al.\cite{FCNT}, and a large number of candidate regions are obtained.
This is a method based on the assumption that candidate area at time $t$ is determined by the estimation result $\mathbf{\hat{x}}_{t-1}=\{\hat{x}_{t-1}, \hat{y}_{t-1}, \hat{w}_{t-1}, \hat{h}_{t-1}\}$ at time $t-1$.
$x_t$ and $y_t$ are sampled from the normal distribution with mean $\hat{x}_{t-1}$,$\hat{y}_{t-1}$ and standard deviation $\sigma_{x,y}\max{(\hat{w}_{t-1},\hat{h}_{t-1})}$, respectively:

\begin{equation}
x_t \sim \mathcal{N}(\hat{x}_{t-1},\sigma_{x,y}\max{(\hat{w}_{t-1},\hat{h}_{t-1})})\label{eq:2-7}
\end{equation}
\begin{equation}
y_t \sim \mathcal{N}(\hat{y}_{t-1},\sigma_{x,y}\max{(\hat{w}_{t-1},\hat{h}_{t-1})})\label{eq:2-8}
\end{equation}

The aspect ratio of $w_t$ and $h_t$ is maintained.
$w_t$ and $h_t$ are sampled by following normal distribution:

\begin{equation}
\frac{w_t}{\hat{w}_{t-1}} = \frac{h_t}{\hat{h}_{t-1}} \sim \mathcal{N}(1,\sigma_{w,h})\label{eq:2-9}
\end{equation}

Based on this sampling method, $N_R$ of candidate regions at time $t$ are generated. From the candidate areas, the best candidate area is selected according to the evaluation method in the next section and used as the prediction area. In this paper, we set $\sigma_{x,y}=0.01$, $\sigma_{w,h}=1/3$, $N_R = 600$.

\subsubsection{Evaluation method of candidate areas}
In case of evaluating the candidate areas on the prediction map,
Wang\cite{FCNT}'s method of ranking the areas by the sum of the values in the candidate area and the Euclidean distance $d$ between the center coordinates at the previous time and the current time are used.
The prediction map takes the values of $0$ or higher for all elements and it is normalized to the range $[0,1]$. Then, a constant $b = 0.2$ is subtracted from all the elements of the prediction map.
The value of a candidate area $R_i,$ where $i\in\{1,2,\ldots,N_R\}$, is defined by following equation:

\begin{equation}
score_i=\sum_{(x,y) \in R_i}(\hat{\mathbf{M}}(x,y)-b)\label{eq:3-9}
\end{equation}
The confidence factor of the candidate area is calculated by this score and the following equation:

\begin{equation}
conf_i=(1-\frac{d_i}{D})(score_i - \min_j{score_j})\label{eq:3-10}
\end{equation}

Where $d_i$ is the Euclidean distance of the $i$ th candidate area and $D$ is half of one side of the ROI.

\subsubsection{Algorithm of FMST}

\renewcommand{\algorithmicrequire}{\textbf{Input:}}
\renewcommand{\algorithmicensure}{\textbf{Output:}}
\begin{algorithm}
\caption{FMST}
\begin{algorithmic}[1]
\REQUIRE Correct area $\mathbf{x}_0$, Images $\mathbf{I}_t, t\in\{0,1,...,T\}$
\ENSURE Prediction area $\mathbf{x}_t$, $t\in\{1, 2, ..., T\}$
\STATE Cut out the ROI from $\mathbf{I}_0$, centering on $\mathbf{x}_0$;
\STATE Generate $\mathbf{M}^l_{\textrm{S}}$ and $\mathbf{M}^l_{\textrm{C}}$;
\STATE Enter the ROI in the CNN and get $\{\mathbf{F}^l_c\}$;
\STATE Get $\mathbf{s}^l_{\textrm{S}}$ and $\mathbf{s}^l_{\textrm{C}}$ by Eq.\ref{eq:2-1};
\STATE $\mathbf{\bar{s}}^l_{\textrm{S}} \leftarrow \mathbf{s}^l_{\textrm{S}}$ , $\mathbf{\bar{s}}^l_{\textrm{C}} \leftarrow \mathbf{s}^l_{\textrm{C}}$;
\STATE Initialize $\mathbf{w}^l_{\textrm{S}}$, $\mathbf{w}^l_{\textrm{C}}$ by Eq.\ref{eq:2-4};
\FOR{$t=1$ to $T$}
\STATE Cut out the ROI from $\mathbf{I}_t$, centering on $\mathbf{x}_{t-1}$;
\STATE Enter the ROI in the CNN and get $\{\mathbf{F}^l_c\}$;
\STATE Obtain $\hat{\mathbf{M}}^l_{\textrm{S}}$, $\hat{\mathbf{M}}^l_{\textrm{C}}$ by Eq.\ref{eq:2-3};
\STATE Normalize $\hat{\mathbf{M}}^l_{\textrm{S}}$, $\hat{\mathbf{M}}^l_{\textrm{C}}$ to range $[0,1]$;
\STATE $\hat{\mathbf{M}}=\sum_{l}(\hat{\mathbf{M}}^l_{\textrm{S}}+\hat{\mathbf{M}}^l_{\textrm{C}})$;
\STATE The region that maximizes Eq.\ref{eq:3-9} on $\hat{\mathbf{M}}$ is defined as the prediction region $\mathbf{x}_{t}$;
\STATE Generate $\mathbf{M}^l_{\textrm{S}}$ and $\mathbf{M}^l_{\textrm{C}}$ using $\mathbf{x}_t$;
\STATE Get $\mathbf{s}^l_{\textrm{S}}$ and $\mathbf{s}^l_{\textrm{C}}$ by Eq.\ref{eq:2-1} using $\{\mathbf{F}^l_c\}$, $\mathbf{M}^l_{\textrm{S}}$ and $\mathbf{M}^l_{\textrm{C}}$;
\STATE Update $\mathbf{\bar{s}}^l_{\textrm{S}}$, $\mathbf{\bar{s}}^l_{\textrm{C}}$ by Eq.\ref{eq:2-5};
\STATE Update $\mathbf{w}^l_{\textrm{S}}$, $\mathbf{w}^l_{\textrm{C}}$ by Eq.\ref{eq:2-4};
\ENDFOR
\end{algorithmic}
\label{alg:tracking}
\end{algorithm}

\subsection{Benchmark for object tracking evaluation}\label{sec2.3}
We use VTB13 in Visual Tracker Benchmark\cite{OTB} as a benchmark.
VTB13 has 50 videos and 51 tracking tasks.
Tracking accuracy is determined by averaging the metrics defined by Wu et al.\cite{OTB} for these 51 tasks.
OPE(One Pass Evaluation) is used as the evaluation method.
OPE is a method giving the correct cordinates of tracking target only in first frame and averaging the results of tracking on the subsequent frames.
In the following section, we use precision plots and success plots as evaluation metrics.
Both metrics visualize the fluctuation of the success rate by continuously changing the permissible error rate.
The precision plots is a metric that evaluates according to the Euclidean distance between the correct answer and the prediction, regardless of the size of the area.
In precision plots, the horizontal axis is the tolerance of the distance between the center coordinates, and the vertical axis is the percentage of frames that fit within the tolerance.
The representative value of precision plots is called precision score, which is the evaluation value with a tolerance 20 pixels.

The success plots is a metric that depends on how the correct area and the predicted area overlap. It is called Intersection over Union(IOU).
IOU is the ratio of the intersection to the union of the two regions.
For success plots, the horizontal axis is the minimum allowable IOU value, and the vertical axis is the percentage of frames that exceed the minimum allowable value.
The success score which is the representative value of success plots is the area under the curve(AUC) drawn by success plots.

\section{FMST using MobileNetV3}\label{sec:3}
In this research, we use Keras 2.3.0 with Tensorflow ver2.0.0.
As hardware, we use Intel Core i7 4790 CPU, 16 GB memory, and NVIDIA GeForce RTX 2080 Ti GPU.

\subsection{Accuracy comparison of FMST which replaced VGG16 with MobileNet}
In this section, we compare the accuracy of FMST which replaced VGG16 with MobileNet V3 Large and Small\cite{2019movilenetv3}.
For MobileNet V3 Large(V3L), we get features from block11, block12 and block 13.
For MobileNet V3 small(V3S), we obtain features from block6, block8 and block 9.
For the benchmark, we use VTB13 introduced in Sec.\ref{sec2.3}.

Table \ref{table:3-1} shows the accuracy comparison of FMST.
In the table, Prec and Succ represent precision score and success score, respectively. 
FPS(C) and FPS(G) indicate the processing speed.
FPS(C) is Flame Per Second(FPS) when the calculation is performed only by the CPU without using the GPU.
FPS(G) is FPS attained by adopting GPU.
The S and C in the model column indicate the types of the applied target maps.
SC means the accuracy of tracking using both $\mathbf{M}^l_{\textrm{S}}$ and $\mathbf{M}^l_{\textrm{C}}$.

\begin{table}[htb]
\centering
\caption{Result of FMST using VGG16 or MobileNetV3}
  \begin{tabular}{lcccc}
    \hline
     Model & Pre(\%) & Suc(\%) & FPS(C) & FPS(G)\\ \hline \hline

    VGG(SC)\cite{yamada} & 81.6 & 56.9 & \textasciitilde 8 & 63 \\ \hline
    VGG(S)\cite{yamada} & 80.0 & 55.2 & \textasciitilde 8 & 63 \\ \hline
    VGG(C)\cite{yamada} & 81.8 & 56.0 & \textasciitilde 8 & 63 \\ \hline \hline
    V3L(SC) & 52.89 & 36.00 & 36.88 & 59.22 \\ \hline
    V3L(S)  & 63.76 & 42.26 & 37.78 & 63.06 \\ \hline
    V3L(C)  & 49.60 & 34.49 & 38.53 & 63.20 \\ \hline \hline
    V3S(SC) & 42.38 & 29.31 & 52.94 & 70.89 \\ \hline
    V3S(S)  & 52.61 & 36.23 & 55.21 & 73.20 \\ \hline
    V3S(C)  & 37.45 & 26.07 & 54.98 & 73.75 \\ \hline
  \end{tabular}
  \label{table:3-1}
\end{table}

Although the method using MobileNet is significantly inferior in accuracy to the original method\cite{yamada} using VGG16, it is advantageous in terms of processing speed.
In an environment that does not use a GPU, the conventional method using VGG16 is around $8$ FPS, whereas the method using MobileNet realizes high-speed processing.
Since the FPS of general video is around 30, both models using V3L and V3S have reached real-time performance even in the environment without GPU.
In addition, when using MobileNet, the accuracy is highest when only the Type S target map is used.

The method using MobileNet has a problem that features appear in a slightly wider range than the actual features on the prediction map.
As a result, the prediction area recursively expanded during the tracking task, and tracking sometimes failed.

To avoid this problem, when the candidate areas are generated by the method in Sec.\ref{chap2:candidate}, the mean size of the areas is resized slightly smaller than the rectangle size of the previous time. The mean size is determined by the sampling based on the following equation:

\begin{equation}
\frac{w_t}{\hat{w}_{t-1}} = \frac{h_t}{\hat{h}_{t-1}} \sim \mathcal{N}(0.996,\sigma_{w,h})\label{eq:3-1}
\end{equation}
The tracking results using this resizing method is shown in Table \ref{table:3-2}.

\begin{table}[htb]
\centering
\caption{Result of FMST with modified area size.}
  \begin{tabular}{lcccc}
    \hline
    Model & Pre(\%) & Suc(\%) & FPS(C) & FPS(G)\\ \hline \hline

    V3L(SC) & 65.25 & 46.15 & 38.07 & 59.89 \\ \hline
    V3L(S) & 72.47 & 50.93 & 39.20 & 61.59 \\ \hline
    V3L(C) & 60.47 & 42.87 & 39.34 & 62.05 \\ \hline \hline

    V3S(SC) & 52.65 & 37.29 & 52.65 & 70.24 \\ \hline
    V3S(S) & 66.13 & 46.73 & 55.68 & 73.04 \\ \hline
    V3S(C) & 45.23 & 33.59 & 55.77 & 74.03 \\ \hline
  \end{tabular}
  \label{table:3-2}
\end{table}

Based on the results mentioned above, only the target map Type S is used hereafter and the size of the candidate area is evaluated by Eq.\ref{eq:3-1}.

\subsection{Ideas to improve the tracking accuracy}\label{sec:3-2}
It was confirmed that high-speed processing can be realized by using mobile CNN. In this section, we will consider improving the tracking accuracy.

\subsubsection{Searching for the optimal ratio}
\begin{figure}[h]
 \begin{minipage}{0.3\hsize}
  \begin{center}
   \includegraphics[height=25mm]{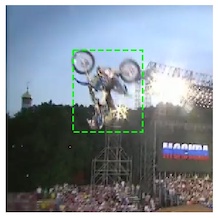}
  \end{center}
  \subcaption[]{ROI.}
  \label{fig:4-vsr-ROI}
 \end{minipage}
 \begin{minipage}{0.3\hsize}
  \begin{center}
   \includegraphics[height=25mm]{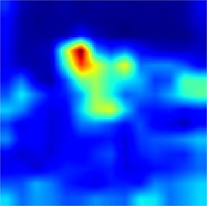}
  \end{center}
  \subcaption[]{top 10 \%.}
  \label{fig:4-10}
 \end{minipage}
 \begin{minipage}{0.3\hsize}
  \begin{center}
   \includegraphics[height=25mm]{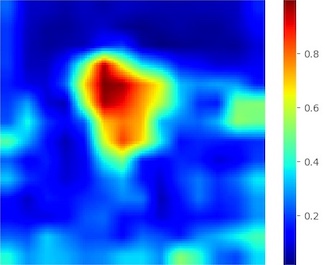}
  \end{center}
  \subcaption[]{top $4.46 \%$.}
  \label{fig:4-446}
 \end{minipage}
 \caption{Examples of prediction maps.}
 \label{fig:4-vsr}
\end{figure}

In the original FMST, the prediction map is generated by summing the feature maps whose the scores are at top 10\%.
However, since the feature maps to be selected for each frame is different, it can be assumed that there is an optimum feature maps selection specific to each frame other than selecting the top 10\%.
Fig.\ref{fig:4-vsr} shows one example to support this assumption.

Fig.\ref{fig:4-vsr-ROI} is input ROI whose green rectangle is the tracking target.
Fig.\ref{fig:4-10} is a prediction map generated by feature maps whose scores is at top $10\%$. Fig.\ref{fig:4-446} is generated by feature maps with top $4.46\%$ score.
Fig.\ref{fig:4-446} is able to extract features of the target more strongly than Fig.\ref{fig:4-10}.

\subsubsection{Using negative features}\label{sec:3_negative}
\begin{figure}[h]
\begin{minipage}{0.3\hsize}
  \begin{center}
  \includegraphics[width=22mm]{fig/chap2/Basketball_ROI.jpg}
  \end{center}
  \subcaption{ROI.}
  \label{negative_roi}
\end{minipage}
\begin{minipage}{0.3\hsize}
  \begin{center}
  \includegraphics[height=23mm]{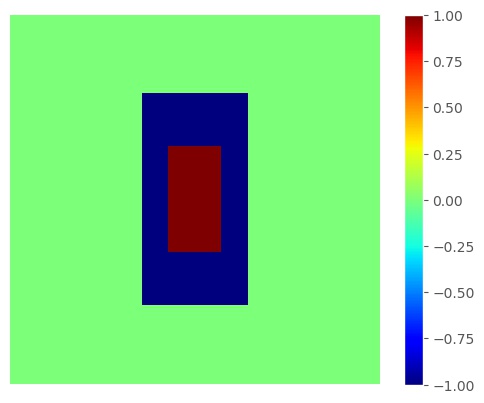}
  \end{center}
  \subcaption{$\mathbf{M}_\mathrm{p}$.}
  \label{positive_tmap}
\end{minipage}
\begin{minipage}{0.3\hsize}
  \begin{center}
  \includegraphics[height=23mm]{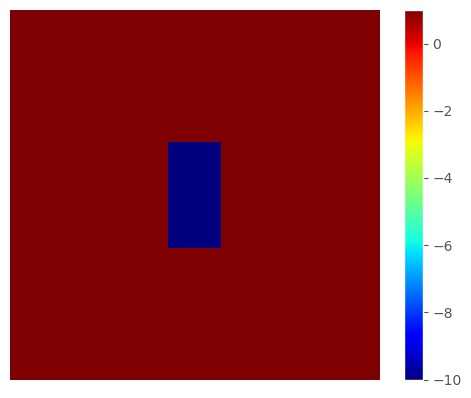}
  \end{center}
  \subcaption{$\mathbf{M}_\mathrm{n}$.}
  \label{negative_tmap}
\end{minipage}
\caption{Examples of ROI, $\mathbf{M}_\mathrm{p}$ and $\mathbf{M}_\mathrm{n}$.}
\label{fig:4_ntmap}
\end{figure}

To remove features other than tracking target, we introduce a negative target map.
Negative target map is a matrix that has negative values for the part corresponding to the tracking target and positive values for other part.
The target map with a positive values for the tracking target is expressed as $\mathbf{M}_\mathrm{p}$, and the negative target map is expressed as $\mathbf{M}_\mathrm{n}$.
Fig.\ref{fig:4_ntmap} shows the examples.
Fig.\ref{negative_roi} represents the input ROI whose green rectangle is the tracking target.
Fig.\ref{positive_tmap} presents a positive target map and Fig.\ref{negative_tmap} depicts a negative target map.

The prediction map generated by $\mathbf{M}_\mathrm{p}$ is referred to as $\hat{\mathbf{M}}_\mathrm{p}$ and the prediction map produced using $\mathbf{M}_\mathrm{n}$ is represented as $\hat{\mathbf{M}}_\mathrm{n}$.

The following equation uses $\hat{\mathbf{M}}_\mathrm{p}$ and $\hat{\mathbf{M}}_\mathrm{n}$ and define the definitive prediction map $\hat{\mathbf{M}}$:

\begin{equation}
\hat{\mathbf{M}}  = \hat{\mathbf{M}}_\mathrm{p} - \alpha\hat{\mathbf{M}}_\mathrm{n}\label{eq:3-2}
\end{equation}

\begin{figure}[h]
 \begin{minipage}{0.3\hsize}
  \begin{center}
   \includegraphics[height=25mm]{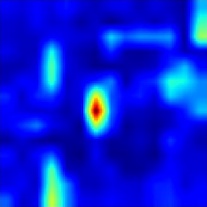}
  \end{center}
  \subcaption{$\hat{\mathbf{M}}_\mathrm{p}$.}
  \label{fig:positive_map}
 \end{minipage}
 \begin{minipage}{0.3\hsize}
  \begin{center}
   \includegraphics[height=25mm]{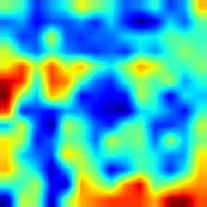}
  \end{center}
  \subcaption{$\hat{\mathbf{M}}_\mathrm{n}$.}
  \label{fig:negative_map}
 \end{minipage}
 \begin{minipage}{0.3\hsize}
  \begin{center}
   \includegraphics[height=25mm]{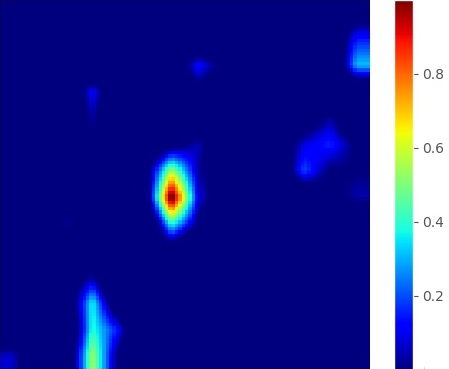}
  \end{center}
  \subcaption{$\hat{\mathbf{M}}(\alpha = 0.5)$.}
  \label{fig:posi_nega}
 \end{minipage}
\caption{Examples of $\hat{\mathbf{M}}_\mathrm{p}$, $\hat{\mathbf{M}}_\mathrm{n}$ and $\hat{\mathbf{M}}$.}
\label{fig:4_negativepmap}
\end{figure}

$\hat{\mathbf{M}}_\mathrm{p}$ and $\hat{\mathbf{M}}_\mathrm{n}$ is normarized to the range $[0,1]$, and the $\alpha$ is a parameter that controls the effect of $\hat{\mathbf{M}}_\mathrm{n}$.
The examples when the input image is Fig.\ref{negative_roi} are shown in Fig.\ref{fig:4_negativepmap}.

Fig.\ref{fig:positive_map} has the strongest activity on the tracking target.
In Fig.\ref{fig:negative_map}, the activity on the tracking target is clearly weak and the activities other than the target is high.
In Fig.\ref{fig:posi_nega}, The definitive prediction map $\hat{\mathbf{M}}$ is able to capture the fratures of the target more powerfully.

\section{Weight vector generation network}
We build a new network based on the improvements in Sec.\ref{sec:3-2}.
The feature maps obtained from the selected layers of MobileNet are combined in the channel direction and regarded as one set of feature maps $\mathbf{F}$.
$\mathbf{F}$ of V3L has $(14\times14\times1824)$ size, and
$\mathbf{F}$ of V3S has $(14\times14\times672)$ size.

\subsection{Architecture of the network}
We build a fully connected network that transforms the score vector $\mathbf{s}$ into a weight vector $\mathbf{w}$.
From this, it can be expected to learn to select the better features for each frames.
By using the ReLU6 as the output activation function, each element of the weight vector has a real value in the range of $[0.0,6.0]$.

We consider the learning flow of the $\mathbf{w}_\mathrm{p}$ generation network for $\hat{\mathbf{M}}_\mathrm{p}$.
First, the input score vector $\mathbf{s}_\mathrm{p}$ to the network is obtained by using the feature map set $\mathbf{F}^{k}$ at time $k$ and the target map $\mathbf{M}^{k}_\mathrm{p}$ based on the correct coordinates at that time.
The element of the score vector $\mathbf{s}_\mathrm{p}$ is obtained from the following equation:
\begin{equation}
 s_{\mathrm{p}c} = \mathrm{sum}(\mathbf{F}^{k}_c \circ \mathbf{M}^{k}_\mathrm{p})\label{eq:5-1}
\end{equation}

We enter $\mathbf{s}_\mathrm{p}$ to the network and get the positive weight vector $\mathbf{w}_\mathrm{p}$.
A prediction map $\hat{\mathbf{M}}_\mathrm{p}$ is generated by using the $\mathbf{w}_\mathrm{p}$ and the feature map set at time $k+1$:

\begin{equation}
 \hat{\mathbf{M}}_\mathrm{p} = \mathbf{F}^{k+1}\mathbf{w}_\mathrm{p} \label{eq:5-2}
\end{equation}

The loss between this $\hat{\mathbf{M}}_\mathrm{p}$ and the target map $\mathbf{M}^{k+1}$ at time $k+1$ is defined by the following equation:

\begin{equation}
 Loss = -\mathrm{sum}(\hat{\mathbf{M}} \circ \mathbf{M}^{k+1})\label{eq:5-3}
\end{equation}

The network learns to reduce this loss.

\begin{figure}[h]
  \centering
  \includegraphics[width=8cm]{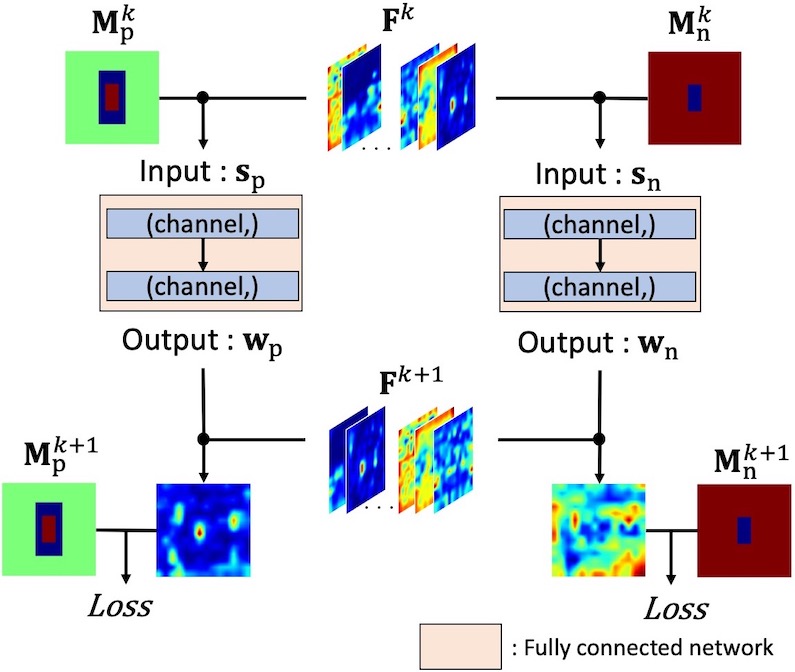}
  \caption[]{$\mathbf{w_p}$ and $\mathbf{w_n}$ generation.}
  \label{fig:chap4_net}
\end{figure}

Since the negative features introduced in Sec.\ref{sec:3_negative} are also used, a network that generates a negative weight vector $\mathbf{w_n}$ is constructed in the same way.
The architecture with positive and negative features is shown in Fig.\ref{fig:chap4_net}

\subsection{Train datasets}
For the traing dataset, we use 49 tracking tasks not included in the VTB13 benchmark.
The dataset is generated by dividing these 49 tasks into frames.
One train data is a pair of an input data of an image with its coordinate information at time $t$ and a teaching data of an image with its coordinate information at time $t+1$.
A total of 29,500 sets are created from 49 tasks, and $10\%$ of them are used as evaluation data.
For learning, Early Stopping with a maximum epoch of $50$ and the patience of $20$ is adopted. The batch size is set to $1$, Adam is used as the optimization method, and the learning rate is set to $0.001$.

\section{Experiments}
\subsection{Tracking accuracy and processing speed}\label{sec:5.1}
\begin{figure}[h]
  \centering
  \includegraphics[width=8cm]{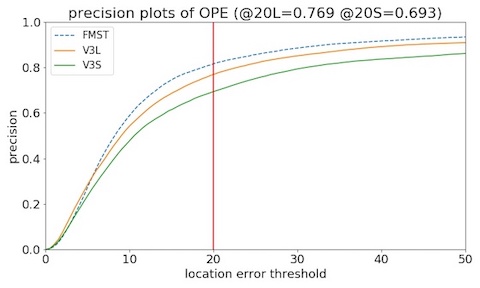}
  \caption[]{Precision plots.}
  \label{fig:chap5_pre}
\end{figure}

\begin{figure}[h]
  \centering
  \includegraphics[width=8cm]{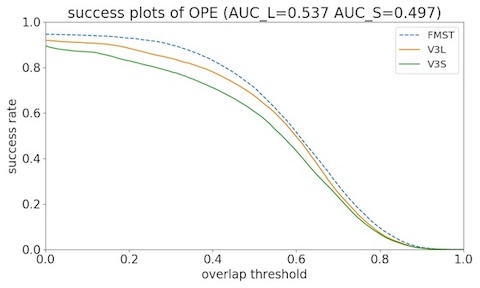}
  \caption[]{Success plots.}
  \label{fig:chap5_suc}
\end{figure}

Table \ref{table:proposed} shows the tracking accuracy of VTB13 in the proposed method. Fig.\ref{fig:chap5_pre} shows the precision plots and Fig.\ref{fig:chap5_suc} represents the success rate. In Fig.\ref{fig:chap5_pre}, the red line indicates the point that is the precision score.
The accuracy of the proposed method is less than that of the original FMST, but improved over the results in Table \ref{table:3-2}.
Moreover, since the processing speed is also improved, it can be seen that the weight vector generation network is not a bottleneck.

\begin{table}[h]
 \centering
 \caption{Comparison of FMST and proposed method.}
 \begin{tabular}{lcccc}
  \hline
  Model & Prec(\%) & Succ(\%) & FPS(C) & FPS(G)\\ \hline \hline
  FMST & 81.56 & 56.86 & \textasciitilde 8& 63 \\
  Ours(V3L) & 76.94 & 53.75 & 41.52 & 64.95 \\
  Ours(V3S) & 69.31 & 49.74 & 60.27 & 77.63 \\
  \hline
 \end{tabular}
 \label{table:proposed}
\end{table}

\subsection{Tracking result and some failure case}
\begin{figure}[h]
 \begin{minipage}{0.33\hsize}
  \begin{center}
   \includegraphics[width=53mm]{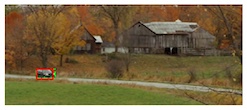}
  \end{center}
 \end{minipage}
 \begin{minipage}{0.33\hsize}
  \begin{center}
   \includegraphics[width=53mm]{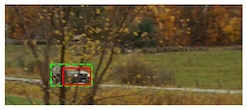}
  \end{center}
 \end{minipage}
 \begin{minipage}{0.33\hsize}
  \begin{center}
   \includegraphics[width=53mm]{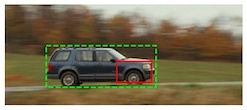}
  \end{center}
 \end{minipage}
 \caption[]{Task CarScale.}
 \label{fig:6.5_CarScale}
\end{figure}

\begin{figure}[h]
 \begin{minipage}{0.33\hsize}
  \begin{center}
   \includegraphics[width=53mm]{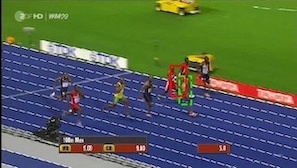}
  \end{center}
 \end{minipage}
 \begin{minipage}{0.33\hsize}
  \begin{center}
   \includegraphics[width=53mm]{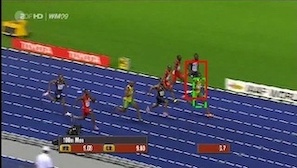}
  \end{center}
 \end{minipage}
 \begin{minipage}{0.33\hsize}
  \begin{center}
   \includegraphics[width=53mm]{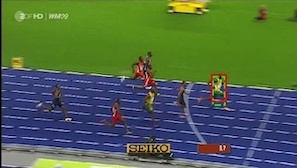}
  \end{center}
 \end{minipage}
 \caption[]{Task Bolt.}
 \label{fig:6.5_Bolt}
\end{figure}

\begin{figure}[h]
 \begin{minipage}{0.33\hsize}
  \begin{center}
   \includegraphics[width=53mm]{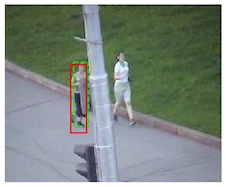}
  \end{center}
 \end{minipage}
 \begin{minipage}{0.33\hsize}
  \begin{center}
   \includegraphics[width=53mm]{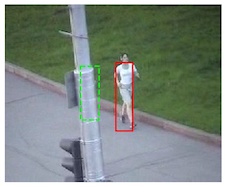}
  \end{center}
 \end{minipage}
 \begin{minipage}{0.33\hsize}
  \begin{center}
   \includegraphics[width=53mm]{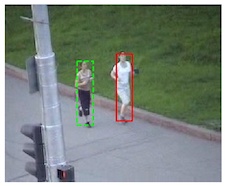}
  \end{center}
 \end{minipage}
 \caption[]{Task Jogging-1.}
 \label{fig:6.5_Jogging}
\end{figure}

\begin{figure}[h!]
 \begin{minipage}{0.33\hsize}
  \begin{center}
   \includegraphics[width=53mm]{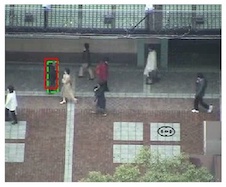}
  \end{center}
 \end{minipage}
 \begin{minipage}{0.33\hsize}
  \begin{center}
   \includegraphics[width=53mm]{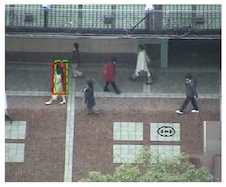}
  \end{center}
 \end{minipage}
 \begin{minipage}{0.33\hsize}
  \begin{center}
   \includegraphics[width=53mm]{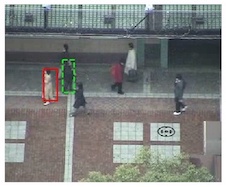}
  \end{center}
 \end{minipage}
 \caption[]{Task Subway.}
 \label{fig:6.5_Subway}
\end{figure}

As shown in Sec.\ref{sec:5.1}, the method introduced in Sec.\ref{sec:3} improves the efficiency of the object tracking. However, some failure cases are found. A part of failure tracking examples is shown in Fig.\ref{fig:6.5_CarScale} to Fig.\ref{fig:6.5_Subway}.
In those images, the green rectangle is the correct area, and the red rectangle indicates the prediction area. 
These examples are the results of the proposal method using V3L.

Fig.\ref{fig:6.5_CarScale} is an example of tracking in the task "CarScale".
It is a task to track a car whose scale changes rapidly.
It can be confirmed that the prediction area is expanding with time, but it has not caught up with the actual scale change.
This is a problem that depends on the standard deviation when generating the candidate region.
By increasing the standard deviation, it becomes possible to respond to sudden scale changes. However, this also causes a problem that the prediction area is unnecessarily expanded. Due to this trade-off, the parameters are set to accommodate gradual scale changes.

Fig.\ref{fig:6.5_Bolt} is an example of tracking in the task "Bolt".
It is a task to track Bolt from among the many runners.
While it can be tracked, it is not clearly distinguishable from other players on the way.
From this, it is considered that this method is not very effective in distinguishing objects of the same category.

Fig.\ref{fig:6.5_Jogging} is an example of tracking in the task "Jogging-1".
One of the two persons is the tracking target.
If the tracked object disappears from the frame because it is hidden by a utility pole, the prediction area will be transferred to the other person.

Fig.\ref{fig:6.5_Subway} is an example of tracking in the task "Subway".
Similar to the example in Fig.\ref{fig:6.5_Bolt}, this is also a failure example that occurs because the objects in the same category cannot be clearly distinguished.

\section{Conclusions}
In this study, we have proposed a tracking method by feature extraction of mobile CNN model. 
When MobileNet was adapted to FMST, high-speed processing was realized, but the tracking accuracy was significantly lower than that of the original one.
Therefore, we constructed a network that automatically selects features.
This automatic feature selection was also performed for negative features, and we succeeded in obtaining performable activity for tracking targets.
The proposed tracking method using this network and the conventional method were compared with tracking accuracy using VTB13.
The precision score of the proposed method was $76.9\%$ and the success score was $53.8\%$, which were inferior to FMST but showed high accuracy.
The processing speed of FNST droped significantly to $8$ FPS in an environment without a GPU.
However, the proposed method achieved a high processing speed of $42$ FPS even in an environment that did not use a GPU.

\bibliography{main}
\bibliographystyle{plain}

\onecolumn

\end{document}